\documentclass[runningheads]{llncs}
\usepackage{graphicx}
\graphicspath{{figures/}}
\usepackage{tabularx}
\usepackage{url}
\usepackage{commath,amsmath,amssymb} 
\usepackage{color}

\begin{document}
\title{UAV-GESTURE: A Dataset for UAV Control and Gesture Recognition} 

\titlerunning{UAV-GESTURE}
%
\author{Asanka G Perera\inst{1}\orcidID{0000-0003-4021-3943} \and
Yee Wei Law\inst{1}\orcidID{0000-0002-5665-0980} \and
Javaan Chahl\inst{1,2}}
%
\authorrunning{A. Perera et al.}
%

\institute{School of Engineering, University of South Australia, Mawson Lakes, SA 5095, Australia\\
\email{asanka.perera@mymail.unisa.edu.au, \{yeewei.law,javaan.chahl\}@unisa.edu.au}\\ \and
Joint and Operations Analysis Division, Defence Science and Technology Group, Melbourne, Victoria 3207, Australia}
\maketitle              
\begin{abstract}
Current UAV-recorded datasets are mostly limited to action recognition and object tracking, whereas the gesture signals datasets were mostly recorded in indoor spaces. Currently, there is no outdoor recorded public video dataset for UAV commanding signals. Gesture signals can be effectively used with UAVs by leveraging the UAVs visual sensors and operational simplicity. To fill this gap and enable research in wider application areas, we present a UAV gesture signals dataset recorded in an outdoor setting. We selected 13 gestures suitable for basic UAV navigation and command from general aircraft handling and helicopter handling signals. We provide 119 high-definition video clips consisting of 37151 frames. The overall baseline gesture recognition performance computed using Pose-based Convolutional Neural Network (P-CNN) is 91.9 \%. All the frames are annotated with body joints and gesture classes in order to extend the dataset's applicability to a wider research area including gesture recognition, action recognition, human pose recognition and situation awareness.

\end{abstract}

\section{Introduction}

Unmanned aerial vehicles (UAVs) can be deployed in a variety of applications such as search and rescue, situational awareness, surveillance and police pursuit by leveraging their mobility and operational simplicity. In some situations, a UAV's ability to recognize the commanding actions of the human operator and to take responsive actions is desirable. Such scenarios might include a firefighter commanding a drone to scan a particular area, a lifeguard directing a drone to monitor a drifting kayaker, or more user-friendly video and photo shooting capabilities. Whether for offline gesture recognition from aerial videos or for equipping UAVs with gesture recognition capabilities, a substantial amount of training data is necessary. However, the majority of the video action recognition datasets consist of ground videos recorded from stationary or dynamic cameras \cite{kang16review}.

Different video datasets recorded from moving and stationary aerial cameras have been published in recent years \cite{kang16review,chaquet13survey}. They have been recorded under different camera and platform settings and have limitations when used with a wide range of human action recognition behaviors demanded today. However, aerial action recognition is still far from perfect. In general, the existing aerial video action datasets are lacking detailed human body shapes to be used with state-of-the-art action recognition algorithms. Many action recognition techniques depend on accurate analysis of human body joints or body frame. It is difficult to use the existing aerial datasets for aerial action or gesture recognition due to one or more of the following reasons: (i) severe perspective distortion -- camera elevation angle closer to $90^\circ$ results in a severely distorted body shape with large head and shoulder, and most of the other body parts being occluded; (ii) the low resolution makes it difficult to retrieve human body and texture details; (iii) motion blur caused by rapid variations of the elevation and pan angles or the movement of the platform; and (iv) camera vibration caused by the engine or the rotors of the UAV.

We introduce a dataset recorded from a low altitude and slow flying mobile platform for gesture recognition. The dataset was created with the intention of capturing full human body details from a relatively low altitude in a way that preserves the maximum detail of the body position. Our dataset is suitable for research involving search and rescue, situational awareness, surveillance, and general action recognition. We assume that in most practical missions, the UAV operator or an autonomous UAV follows these general rules: (i) it does not fly so low that it poses danger to the civilians, ground-based structures, or itself; (ii) it does not fly so high or so fast that it loses too much detail in the images it captures; (iii) it hovers to capture the details of an interesting scene; and (iv) it records human subjects from a viewpoint that causes minimum perspective distortion and maximum body details. Our dataset was created following these guidelines to represent 13 command gesture classes. The gestures were selected from general aircraft handling and helicopter handling signals \cite{navy97aircraft}. All the videos were recorded at high-definition (HD) resolution, enabling the gesture videos to be used in general gesture recognition and gesture-based autonomous system control research. To our knowledge, this is the first dataset presenting gestures captured from a moving aerial camera in an outdoor setting.

\section{Related work}

A complete list and description of recently published action recognition datasets is available in \cite{kang16review,chaquet13survey}, and gesture recognition datasets can be found in \cite{simon14survey,pramod15recent}. Here, we discuss some selected studies related to our work. 

Detecting human action from an aerial view is more challenging than from a fronto-parallel view. Created by Oh et al.~\cite{oh11large}, the large-scale VIRAT dataset contains about 550 videos, recorded from static and moving cameras covering 23 event types over 29 hours. The VIRAT ground dataset has been recorded from stationary aerial cameras (e.g., overhead mounted surveillance cameras) at multiple locations with resolutions of 1080$\times$1920 and 720$\times$1280. Both aerial and ground-based datasets have been recorded in uncontrolled and cluttered backgrounds. However, in the VIRAT aerial dataset, the low resolution of 480$\times$720 precludes retrieval of rich activity information from relatively small human subjects.

A 4K-resolution video dataset called Okutama-Action was introduced in \cite{barekatain17okutama} for concurrent action detection by multiple subjects. The videos have been recorded in a relatively clutter-free baseball field using 2 UAVs. There are 12 actions under abrupt camera movements, altitudes from 10 to 45 meters and different view angles. The camera elevation angle of 90 degrees causes a severe distortion in perspective and self-occlusions in videos. 

Other notable aerial action datasets are UCF aerial action \cite{online11ucfaerial}, UCF-ARG \cite{online11ucfarg} and Mini-drone \cite{bonetto15privacy}. UCF aerial action and UCF ARG have been recorded using an R/C-controlled blimp and a helium balloon respectively. Both datasets contain similar action classes. However, UCF aerial action is a single-view dataset while UCF ARG is a multi-view dataset recorded from aerial, rooftop and ground cameras. The Mini-drone dataset has been developed as a surveillance dataset to evaluate different aspects and definitions of privacy. This dataset was recorded in a car park using a drone flying at a low altitude and the actions are categorized as normal, suspicious and illicit behaviors. 

Gesture recognition has been studied extensively in recent years \cite{pramod15recent,simon14survey}. However, the gesture-based UAV control studies available in the literature are mostly limited to indoor environments or static gestures \cite{lee18forecasting,costante14personalizing,pfeil13exploring}, restricting their applicability to real-world scenarios. The datasets used for these works were mostly recorded indoors using RGB-D images \cite{guyon14chalearn,ruffieux13chairgest,shahroudy16ntu} or RGB images \cite{zhe09recognizing,carol12challenges}. An aircraft handling signal dataset similar to ours in terms of gesture classes is available in \cite{song11tracking}. It has been created using VICON cameras and a stereo camera with a static indoor background. However, these gesture datasets cannot be used in aerial gesture studies. We selected some gesture classes from \cite{song11tracking} when creating our dataset.

\section{Preparing the dataset}

This section discusses the collection process of the dataset, the types of gestures recorded in the dataset, and the usefulness of the dataset for vision-related research purposes.

\subsection{Data collection}

The data was collected on an unsettled road located in the middle of a wheat field from a rotorcraft UAV (3DR Solo) in slow and low-altitude flight. For video recording, we used a GoPro Hero 4 Black camera with an anti-fish eye replacement lens (5.4mm, 10MP, IR CUT) and a 3-axis Solo gimbal. We provide the videos with HD ($1920\times1080$) formats at 25 fps. The gestures were recorded on two separate days. The participants were asked to perform the gestures in a selected section of the road. A total of 13 gestures have been recorded while the UAV was hovering in front of the subject. In these videos, the subject is roughly in the middle of the frame and performs each gesture five to ten times. 

When recording the gestures, sometimes the UAV drifts from its initial hovering position due to wind gusts. This adds random camera motion to the videos making them closer to practical scenarios.

\subsection{Gesture selection}

\begin{figure}
\centering
\includegraphics[width=\textwidth]{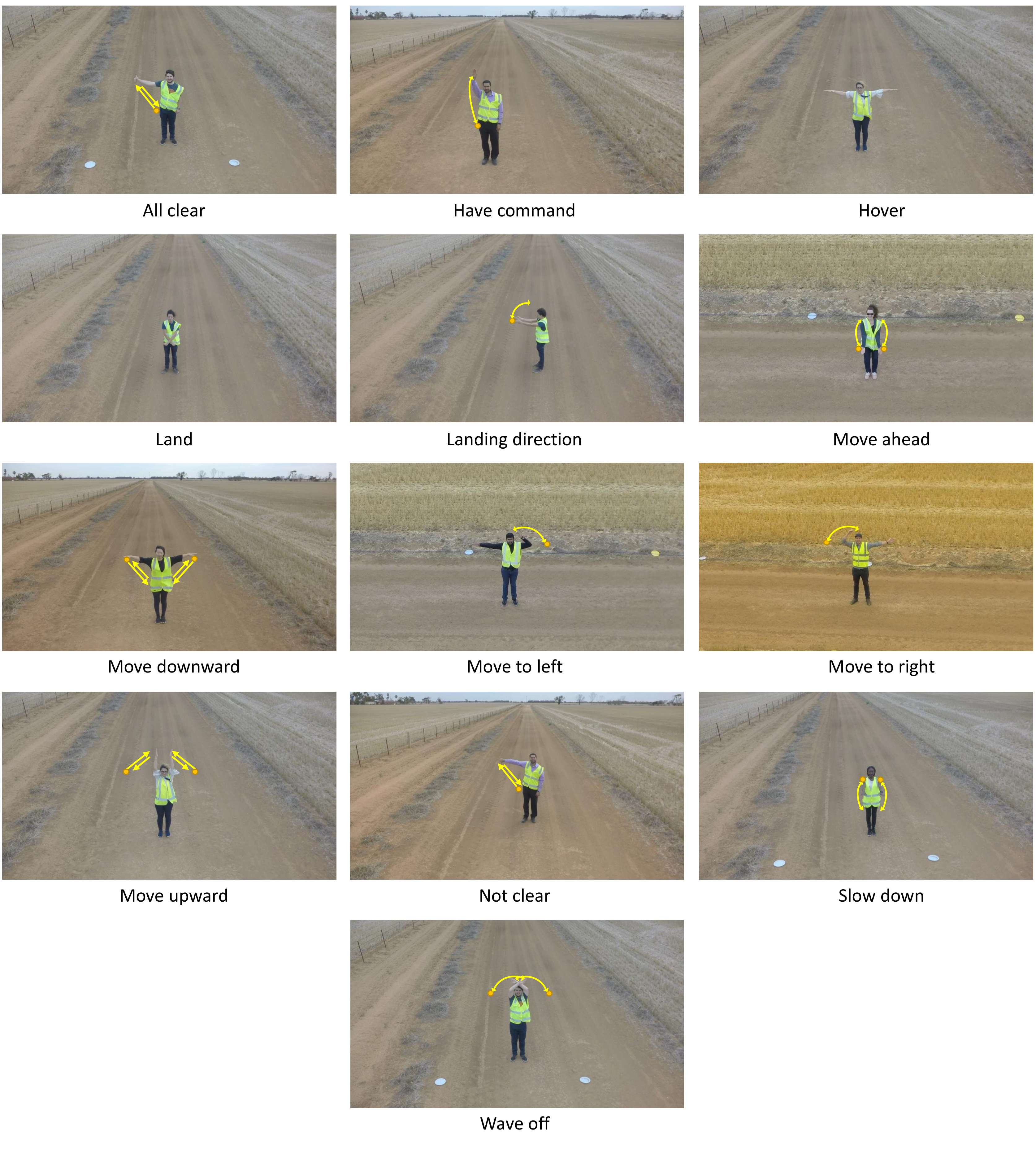}
\caption{The selected thirteen gestures are shown with one selected image from each gesture. The arrows indicate the hand movement directions. The amber color markers roughly designate the start and end positions of the palm for one repetition. The \emph{Hover} and \emph{Land} gestures are static gestures.}
\label{fig:gestures}
\end{figure}

The gestures were selected from general aircraft handling signals and helicopter handling signals available in the Aircraft Signals NATOPS manual \cite[Ch. 2--3]{navy97aircraft}. The selected 13 gestures are shown in Fig.~\ref{fig:gestures}. When selecting the gestures, we avoided aircraft and helicopter specific gestures. The gestures were selected to meet the following criteria: (i) they should be easily identifiable from a moving platform, (ii) the gestures need to be crisp enough to be differentiated from each another, (iii) they need to be simple enough to be repeated by an untrained individual, (iv) the gestures should be applicable to basic UAV navigation control, and (v) the selected gestures should be a mixture of static and dynamic gestures to enable other possible applications such as taking ``selfies''.

\subsection{Variations in data}
  
The actors that participated in this dataset are not professionals in aircraft handling signals. They were shown how to do a particular gesture by another person who was standing in front of them, and then asked to do the same towards the UAV. Therefore, each actor performed the gestures slightly differently. There are rich variations in the recorded gestures in terms of the phase, orientation, camera movement and the body shape of the actors. In some videos, the skin color of the actor is close to the background color. These variations create a challenging dataset for gesture recognition, and also makes it more representative of real-world situations.

The dataset was recorded on two separate days and involved a total of eight participants. Two participants performed the same gestures on both days. For a particular gesture performed by a participant in the two settings, the two videos have significant differences in the background, clothing, camera to subject distance and natural variations in hand movements. Due to these visual variations in the dataset, we consider the total number of actors to be 10.

\subsection{Dataset annotations}

\begin{figure}
\centering
\includegraphics[width=\textwidth]{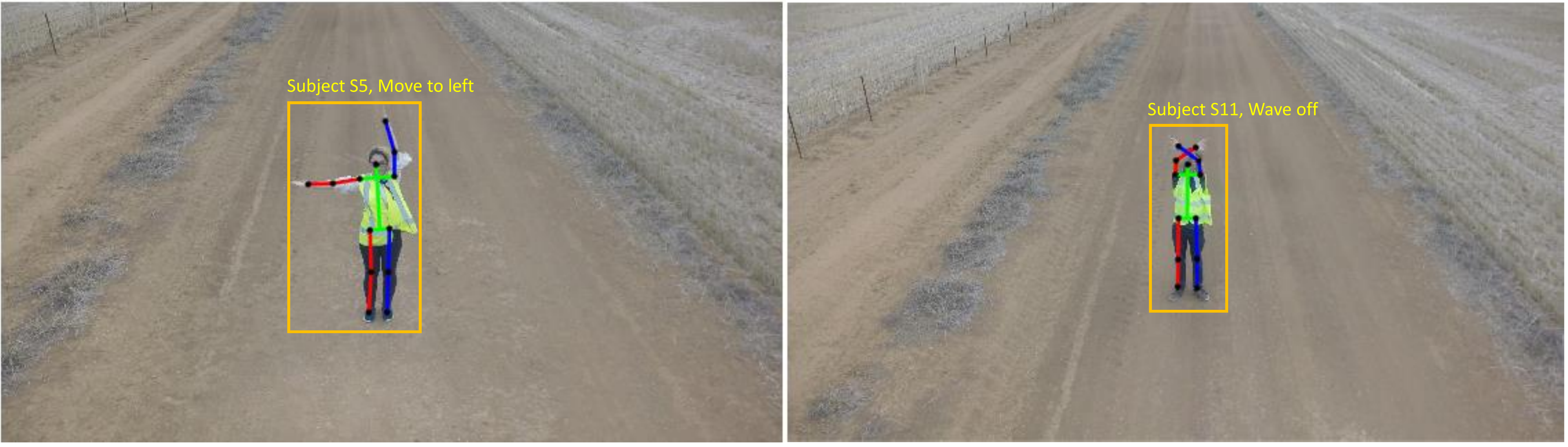}
\caption{Examples of body joint annotations. Image on the left is from the \emph{Move to left} class, whereas the image on the right is from the \emph{Wave off} class.}
\label{fig:annotation}
\end{figure}

We used an extended version of online video annotation tool VATIC~\cite{vondrick13efficiently} to annotate the videos. Thirteen body joints are annotated in 37151 frames, namely ankles, knees, hip-joint, wrists, elbows, shoulders and head. Two annotated images are shown in Figure~\ref{fig:annotation}. Each annotation also comes with the gesture class, subject identity and bounding box. The bounding box is created by adding a margin to the minimum and maximum coordinates of joint annotations in both $x$ and $y$ directions.

\subsection{Dataset summary}

\begin{figure}
\centering
\includegraphics[width=\textwidth]{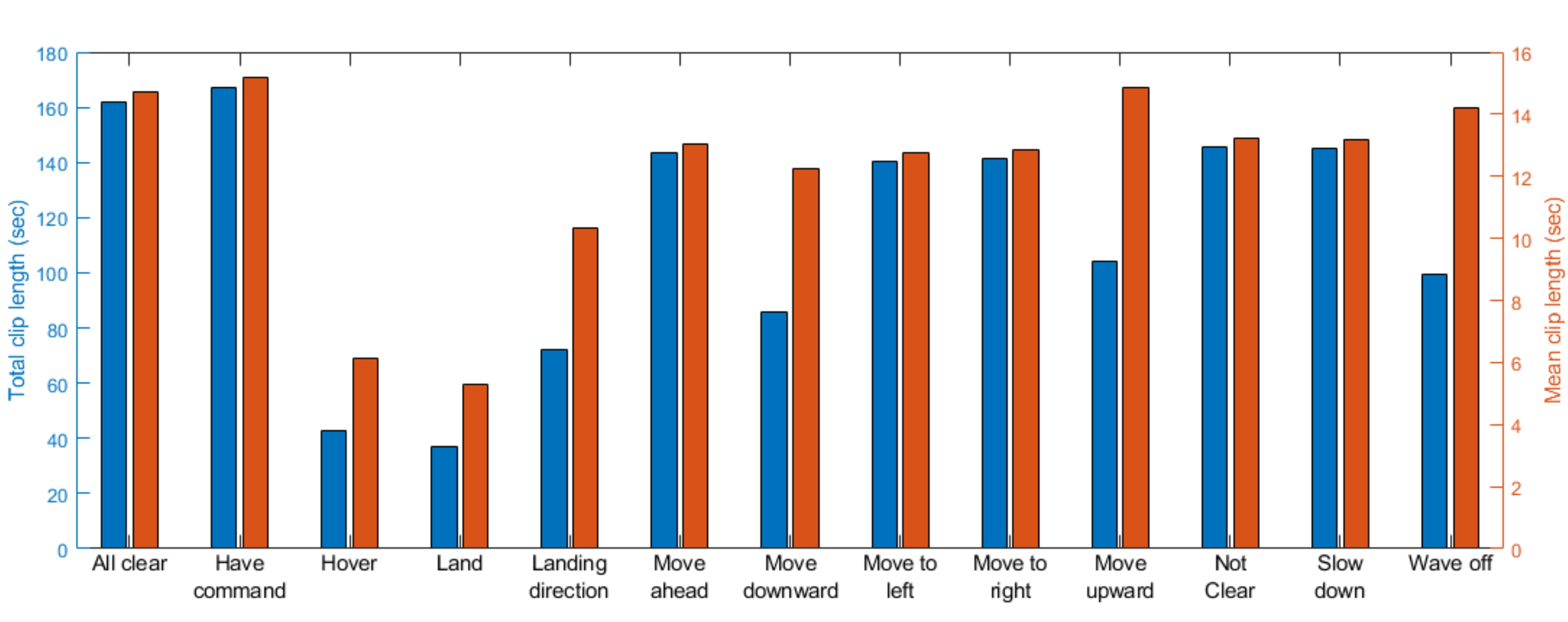}
\caption{The total clip length (blue) and the mean clip length (amber) are shown in the same graph in seconds. Note the former is one order of magnitude higher than the latter.}
\label{fig:total_and_mean_lengths}
\end{figure}

\begin{table}
	\centering
	\caption{A summary of the dataset.}\label{table:summary}
	\begin{tabular}{|l|l|}
	\hline
	Feature & Value\\
	\hline
	\# Gestures           &   13\\
	\# Actors             &   10\\
	\# Clips              &   119\\
	\# Clips per class    &   7-11\\
	Repetitions per class &   5-10\\
	Mean clip length      &   12.5 sec\\
	Total duration        &   24.76 mins\\
	Min clip length       &   3.6 sec\\
	Max clip length       &   23.44 sec\\
	\# Frames             &   37151\\
	Frame rate            &   25 fps\\
	Resolution            &   1920$\times$1080\\
	Camera motion         &   Yes, slight\\
	Annotation            &   Bounding box, body joints\\
	\hline
	\end{tabular}
\end{table}

\setlength{\tabcolsep}{4pt}
	\begin{table}
	\centering
	\caption{Comparison with recently published video datasets.}\label{table:compare}
	\scriptsize
	\begin{tabularx}{\textwidth}{|X|X|X|l|l|l|l|l|}
	\hline
	Dataset & Scenario & Purpose & Environment & Frames & Classes & Resolution & Year\\
	\hline
	UT Interaction \cite{ryoo09spation}       &   Surveillance        &   Action recognition    &   Outdoor &   36k   &   6   &   360$\times$240    &   2010\\
	NATOPS \cite{song11tracking}              &   Aircraft signaling  &   Gesture recognition   &   Indoor  &   N/A   &   24  &   320$\times$240    &   2011\\
	VIRAT \cite{oh11large}                    &   Drone, surveillance &   Event recognition     &   Outdoor &   Many  &   23  &   Varying           &   2011\\
	UCF101 \cite{soomro12ucf101}              &   YouTube             &   Action recognition    &   Varying &   558k  &   24  &   320$\times$240    &   2012\\
	J-HMDB \cite{jhuang13towards}             &   Movies, YouTube     &   Action recognition    &   Varying &   32k   &   21  &   320$\times$240    &   2013\\
	Mini-drone \cite{bonetto15privacy}        &   Drone               &   Privacy protection    &   Outdoor &   23.3  &   3   &   1920$\times$1080  &   2015\\
	Campus \cite{robicquet16learning}         &   Surveillance        &   Object tracking       &   Outdoor &   11.2k &   1   &   1414$\times$2019  &   2016\\
	Okutama-Action~\cite{barekatain17okutama} &   Drone               &   Action recognition    &   Outdoor &   70k   &   13  &   3840$\times$2160  &   2017\\
	UAV-GESTURE                               &   Drone               &   Gesture recognition   &   Outdoor &   37.2k &   13  &   1920$\times$1080  &   2018\\
	\hline
	\end{tabularx}
	\normalsize
\end{table}
\setlength{\tabcolsep}{4pt}

The dataset contains a total of 37151 frames distributed over 119, 25 fps, 1920$\times$1080 video clips. All the frames are annotated with the gesture classes and body joints. There are 10 actors in the dataset, and they perform 5-10 repetitions of each gesture. Each gesture lasts about 12.5 sec on average. A summary of the dataset is given in Table~\ref{table:summary}. The total clip length (blue bars) and mean clip length (amber bars) for each class are shown in Figure~\ref{fig:total_and_mean_lengths}. 

In Table~\ref{table:compare}, we compare our dataset with eight recently published video datasets. These datasets have helped to progress research in action recognition, gesture recognition, event recognition and object tracking. The closest dataset in terms of the class types and the purpose is the NATOPS aircraft signals dataset that was created using 24 selected gestures.

\section{Experimental Results}

\begin{figure}
\centering
\includegraphics[width=\textwidth]{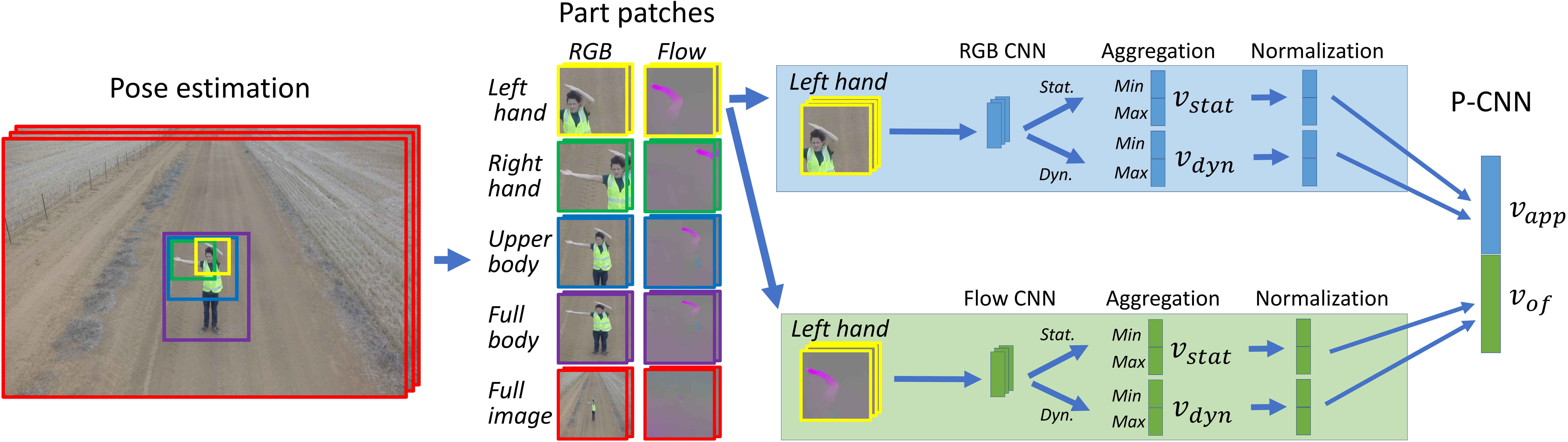}
\caption{The P-CNN feature descriptor \cite{cheron15pcnn}: The steps shown in the diagram correspond to an example P-CNN computation for body part \emph{left hand}.
}
\label{fig:pcnn}
\end{figure}

We performed an experiment on the dataset using Pose-based Convolutional Neural Network (P-CNN) descriptors \cite{cheron15pcnn}. A P-CNN descriptor aggregates motion and appearance information along tracks of human body parts (right hand, left hand, upper body and full body). The P-CNN descriptor was originally introduced for action recognition. Since our dataset contains gestures with full body poses, P-CNN is also a suitable method for full-body gesture recognition. In P-CNN, the body-part patches of the input image are extracted using the human pose and corresponding body parts. For body joint estimation, we used the state-of-the-art OpenPose \cite{cao17realtime} pose estimator which is an extension of Convolutional Pose Machines \cite{wei16convolutional}. Similar to the original P-CNN implementation, the optical flow for each consecutive pair of images was computed using Brox et al.'s method~\cite{brox04high}. 

A diagram showing P-CNN feature extraction is given in Figure \ref{fig:pcnn}. For each body part and full image, the appearance (RGB) and optical flow patches are extracted and their CNN features are computed using two pre-trained networks. For appearance patches, the publicly available ``VGG-f'' network \cite{chatfield14return} is used, whereas for optical flow patches, the motion network from Gkioxari and Malik's Action Tube implementation \cite{gkioxari15finding} is used. Static and dynamic features are separately aggregated over time to obtain a static video descriptor $v_{stat}$ and a dynamic video descriptor $v_{stat}$ respectively. The static features are the (i) distances between body joints, (ii) orientations of the vectors connecting pairs of joints, and (iii) inner angles spanned by vectors connecting all triplets of joints. The dynamic features are computed from trajectories of body joints. We select the \emph{Min} and \emph{Max} aggregation schemes, because of their high accuracies over other schemes when used with P-CNN \cite{cheron15pcnn} on the JHMDB dataset \cite{jhuang13towards} for action recognition. The \emph{Min} and \emph{Max} aggregation schemes compute the minimum and maximum values respectively for each descriptor dimension over all video frames. The static and dynamic video descriptors can be defined as
\begin{align}
v_{stat} &= \sbr{m_1, \ldots , m_k, M_1, \ldots, M_k}^\top, \\
v_{dyn}  &= \sbr{\Delta m_1, \ldots , \Delta m_k, \Delta M_1, \ldots, \Delta M_k}^\top,
\end{align}
where, $m$ and $M$ correspond to the minimum and maximum values for each video descriptor dimension $1,\ldots,k$. $\Delta$ represents temporal differences in the video descriptors. The aggregated features ($v_{stat}$ and $v_{dyn}$) are normalized and concatenated over the number of body parts to obtain appearance features $v_{app}$ and flow features $v_{of}$. The final P-CNN descriptor is obtained by concatenating $v_{app}$ and $v_{of}$.

 The evaluation metric selected for the experiment is accuracy. Accuracy is calculated using the scores returned by the action classifiers. There are three training and testing splits for UAV-GESTURE dataset. In Table \ref{table:pcnn_datasets}, the mean accuracy is compared with the evaluation results reported in \cite{cheron15pcnn} for the JHMDB \cite{jhuang13towards} and MPII Cooking \cite{rohrbach12database} datasets. For the JHMDB and MPII Cooking datasets, the poses are estimated using the pose estimator described in \cite{cherian14mixing}. However, we use OpenPose \cite{cao17realtime} for UAV-GESTURE, because OpenPose has been used as the body joint detector in notable pose-based action recognition studies \cite{girdhar17attentional,sijie18spatial,piergiovanni18fine}, and has reportedly the best performance \cite{cao17realtime}.

\begin{table}
	\centering
	\caption{The best reported P-CNN action recognition results for different datasets.}\label{table:pcnn_datasets}
	\begin{tabular}{|l|l|l|}
	\hline
	Dataset & Remarks & Accuracy(\%)\\
	\hline
	JHMDB                     &   Res: 320 $\times$ 240, Pose estimation: \cite{cherian14mixing}     &   74.2\\
	MPII Cooking              &   Res: 1624 $\times$ 1224, Pose estimation: \cite{cherian14mixing}   &   62.3\\
	UAV-GESTURE               &   Res: 1920 $\times$ 1080, Pose estimation: OpenPose \cite{cao17realtime}   &   91.9\\
	\hline
	\end{tabular}
\end{table}

\section{Conclusion}

We presented a gesture dataset recorded by a hovering UAV. The dataset contains 119 HD videos lasting a total of 24.78 minutes. The dataset was prepared using 13 selected gestures from the set of general aircraft handling and helicopter handling signals. The gestures were recorded  from 10 participants in an outdoor setting. The rich variation of body size, camera motion, and phase, makes our dataset challenging for gesture recognition. The dataset is annotated for human body joints and action classes to extend its applicability to a wider research community. We evaluated this new dataset using P-CNN descriptors and reported an overall baseline action recognition accuracy of 91.9 \%. This dataset is useful for research involving gesture-based unmanned aerial vehicle or unmanned ground vehicle control, situation awareness, general gesture recognition, and general action recognition. The UAV-GESTURE dataset is available at \url{https://github.com/asankagp/UAV-GESTURE}.

\section*{Acknowledgement}
This project was partly supported by Project Tyche, the Trusted Autonomy Initiative of the Defence Science and Technology Group (grant number myIP6780).

\bibliographystyle{splncs04}
\bibliography{ref}

\begin{thebibliography}{10}
\providecommand{\url}[1]{\texttt{#1}}
\providecommand{\urlprefix}{URL }
\providecommand{\doi}[1]{https://doi.org/#1}

\bibitem{barekatain17okutama}
Barekatain, M., Martí, M., Shih, H.F., Murray, S., Nakayama, K., Matsuo, Y.,
  Prendinger, H.: Okutama-action: An aerial view video dataset for concurrent
  human action detection. In: 2017 IEEE Conference on Computer Vision and
  Pattern Recognition Workshops (CVPRW). pp. 2153--2160 (July 2017).
  \doi{10.1109/CVPRW.2017.267}

\bibitem{bonetto15privacy}
Bonetto, M., Korshunov, P., Ramponi, G., Ebrahimi, T.: Privacy in mini-drone
  based video surveillance. In: 2015 11th IEEE International Conference and
  Workshops on Automatic Face and Gesture Recognition (FG). vol.~04, pp.~1--6
  (May 2015). \doi{10.1109/FG.2015.7285023}

\bibitem{brox04high}
Brox, T., Bruhn, A., Papenberg, N., Weickert, J.: High accuracy optical flow
  estimation based on a theory for warping. In: Pajdla, T., Matas, J. (eds.)
  Computer Vision - ECCV 2004. pp. 25--36. Springer Berlin Heidelberg, Berlin,
  Heidelberg (2004)

\bibitem{cao17realtime}
Cao, Z., Simon, T., Wei, S.E., Sheikh, Y.: Realtime multi-person 2d pose
  estimation using part affinity fields. In: CVPR (2017)

\bibitem{carol12challenges}
Carol~Neidle, A.T., Sclaroff, S.: 5th workshop on the representation and
  processing of sign languages: Interactions between corpus and lexicon (May
  2012)

\bibitem{chaquet13survey}
Chaquet, J.M., Carmona, E.J., Fernández-Caballero, A.: A survey of video
  datasets for human action and activity recognition. Computer Vision and Image
  Understanding  \textbf{117}(6),  633 -- 659 (2013).
  \doi{https://doi.org/10.1016/j.cviu.2013.01.013},
  \url{http://www.sciencedirect.com/science/article/pii/S1077314213000295}

\bibitem{chatfield14return}
Chatfield, K., Simonyan, K., Vedaldi, A., Zisserman, A.: Return of the devil in
  the details: Delving deep into convolutional nets. CoRR
  \textbf{abs/1405.3531} (2014), \url{http://arxiv.org/abs/1405.3531}

\bibitem{cherian14mixing}
Cherian, A., Mairal, J., Alahari, K., Schmid, C.: Mixing body-part sequences
  for human pose estimation. In: The IEEE Conference on Computer Vision and
  Pattern Recognition (CVPR) (June 2014)

\bibitem{cheron15pcnn}
Cheron, G., Laptev, I., Schmid, C.: P-cnn: Pose-based cnn features for action
  recognition. In: The IEEE International Conference on Computer Vision (ICCV)
  (December 2015)

\bibitem{costante14personalizing}
Costante, G., Bellocchio, E., Valigi, P., Ricci, E.: Personalizing vision-based
  gestural interfaces for hri with uavs: a transfer learning approach. In: 2014
  IEEE/RSJ International Conference on Intelligent Robots and Systems. pp.
  3319--3326 (Sept 2014). \doi{10.1109/IROS.2014.6943024}

\bibitem{girdhar17attentional}
Girdhar, R., Ramanan, D.: Attentional pooling for action recognition. In:
  Guyon, I., Luxburg, U.V., Bengio, S., Wallach, H., Fergus, R., Vishwanathan,
  S., Garnett, R. (eds.) Advances in Neural Information Processing Systems 30,
  pp. 34--45. Curran Associates, Inc. (2017),
  \url{http://papers.nips.cc/paper/6609-attentional-pooling-for-action-recognition.pdf}

\bibitem{gkioxari15finding}
Gkioxari, G., Malik, J.: Finding action tubes. In: The IEEE Conference on
  Computer Vision and Pattern Recognition (CVPR) (June 2015)

\bibitem{guyon14chalearn}
Guyon, I., Athitsos, V., Jangyodsuk, P., Escalante, H.J.: The chalearn gesture
  dataset (cgd 2011). Machine Vision and Applications  \textbf{25}(8),
  1929--1951 (Nov 2014)

\bibitem{jhuang13towards}
Jhuang, H., Gall, J., Zuffi, S., Schmid, C., Black, M.J.: Towards understanding
  action recognition. In: 2013 IEEE International Conference on Computer
  Vision. pp. 3192--3199 (Dec 2013). \doi{10.1109/ICCV.2013.396}

\bibitem{kang16review}
Kang, S., Wildes, R.P.: Review of action recognition and detection methods.
  CoRR  \textbf{abs/1610.06906} (2016), \url{http://arxiv.org/abs/1610.06906}

\bibitem{lee18forecasting}
Lee, J., Tan, H., Crandall, D., \v{S}abanovi\'{c}, S.: Forecasting hand
  gestures for human-drone interaction. In: Companion of the 2018 ACM/IEEE
  International Conference on Human-Robot Interaction. pp. 167--168. HRI '18,
  ACM, New York, NY, USA (2018). \doi{10.1145/3173386.3176967},
  \url{http://doi.acm.org/10.1145/3173386.3176967}

\bibitem{zhe09recognizing}
Lin, Z., Jiang, Z., Davis, L.S.: Recognizing actions by shape-motion prototype
  trees. In: 2009 IEEE 12th International Conference on Computer Vision. pp.
  444--451 (Sept 2009). \doi{10.1109/ICCV.2009.5459184}

\bibitem{oh11large}
Oh, S., Hoogs, A., Perera, A., Cuntoor, N., Chen, C.C., Lee, J.T., Mukherjee,
  S., Aggarwal, J.K., Lee, H., Davis, L., Swears, E., Wang, X., Ji, Q., Reddy,
  K., Shah, M., Vondrick, C., Pirsiavash, H., Ramanan, D., Yuen, J., Torralba,
  A., Song, B., Fong, A., Roy-Chowdhury, A., Desai, M.: A large-scale benchmark
  dataset for event recognition in surveillance video. In: CVPR 2011. pp.
  3153--3160 (June 2011). \doi{10.1109/CVPR.2011.5995586}

\bibitem{pfeil13exploring}
Pfeil, K., Koh, S.L., LaViola, J.: Exploring 3d gesture metaphors for
  interaction with unmanned aerial vehicles. In: Proceedings of the 2013
  International Conference on Intelligent User Interfaces. pp. 257--266. IUI
  '13, ACM, New York, NY, USA (2013). \doi{10.1145/2449396.2449429},
  \url{http://doi.acm.org/10.1145/2449396.2449429}

\bibitem{piergiovanni18fine}
Piergiovanni, A.J., Ryoo, M.S.: Fine-grained activity recognition in baseball
  videos. CoRR  \textbf{abs/1804.03247} (2018),
  \url{http://arxiv.org/abs/1804.03247}

\bibitem{pramod15recent}
Pisharady, P.K., Saerbeck, M.: Recent methods and databases in vision-based
  hand gesture recognition: A review. Computer Vision and Image Understanding
  \textbf{141},  152 -- 165 (2015).
  \doi{https://doi.org/10.1016/j.cviu.2015.08.004},
  \url{http://www.sciencedirect.com/science/article/pii/S1077314215001794}

\bibitem{robicquet16learning}
Robicquet, A., Sadeghian, A., Alahi, A., Savarese, S.: Learning social
  etiquette: Human trajectory understanding in crowded scenes. In: Leibe, B.,
  Matas, J., Sebe, N., Welling, M. (eds.) Computer Vision -- ECCV 2016. pp.
  549--565. Springer International Publishing, Cham (2016)

\bibitem{rohrbach12database}
Rohrbach, M., Amin, S., Andriluka, M., Schiele, B.: A database for fine grained
  activity detection of cooking activities. In: 2012 IEEE Conference on
  Computer Vision and Pattern Recognition. pp. 1194--1201 (June 2012).
  \doi{10.1109/CVPR.2012.6247801}

\bibitem{ruffieux13chairgest}
Ruffieux, S., Lalanne, D., Mugellini, E.: Chairgest: A challenge for multimodal
  mid-air gesture recognition for close hci. In: Proceedings of the 15th ACM on
  International Conference on Multimodal Interaction. pp. 483--488. ICMI '13,
  ACM, New York, NY, USA (2013). \doi{10.1145/2522848.2532590},
  \url{http://doi.acm.org/10.1145/2522848.2532590}

\bibitem{simon14survey}
Ruffieux, S., Lalanne, D., Mugellini, E., Abou~Khaled, O.: A survey of datasets
  for human gesture recognition. In: Kurosu, M. (ed.) Human-Computer
  Interaction. Advanced Interaction Modalities and Techniques. pp. 337--348.
  Springer International Publishing, Cham (2014)

\bibitem{ryoo09spation}
Ryoo, M.S., Aggarwal, J.K.: Spatio-temporal relationship match: Video structure
  comparison for recognition of complex human activities. In: 2009 IEEE 12th
  International Conference on Computer Vision. pp. 1593--1600 (Sept 2009).
  \doi{10.1109/ICCV.2009.5459361}

\bibitem{shahroudy16ntu}
Shahroudy, A., Liu, J., Ng, T.T., Wang, G.: Ntu rgb+d: A large scale dataset
  for 3d human activity analysis. In: The IEEE Conference on Computer Vision
  and Pattern Recognition (CVPR) (June 2016)

\bibitem{song11tracking}
Song, Y., Demirdjian, D., Davis, R.: Tracking body and hands for gesture
  recognition: Natops aircraft handling signals database. In: Face and Gesture
  2011. pp. 500--506 (March 2011). \doi{10.1109/FG.2011.5771448}

\bibitem{soomro12ucf101}
Soomro, K., Zamir, A.R., Shah, M.: {UCF101:} {A} dataset of 101 human actions
  classes from videos in the wild. Tech. rep., UCF Center for Research in
  Computer Vision (2012)

\bibitem{online11ucfaerial}
{University of Central Florida}: {UCF aerial action dataset}.
  \url{http://crcv.ucf.edu/data/UCF_Aerial_Action.php} (November 2011)

\bibitem{online11ucfarg}
{University of Central Florida}: {UCF-ARG Data Set}.
  \url{http://crcv.ucf.edu/data/UCF-ARG.php} (November 2011)

\bibitem{navy97aircraft}
{U.S. Navy}: Aircraft signals natops manual, navair 00-80t-113  (1997),
  \url{http://www.navybmr.com/study%20material/NAVAIR_113.pdf}

\bibitem{vondrick13efficiently}
Vondrick, C., Patterson, D., Ramanan, D.: Efficiently scaling up crowdsourced
  video annotation. International Journal of Computer Vision  \textbf{101}(1),
  184--204 (Jan 2013). \doi{10.1007/s11263-012-0564-1}

\bibitem{wei16convolutional}
Wei, S.E., Ramakrishna, V., Kanade, T., Sheikh, Y.: Convolutional pose
  machines. In: The IEEE Conference on Computer Vision and Pattern Recognition
  (CVPR) (June 2016)

\bibitem{sijie18spatial}
Yan, S., Xiong, Y., Lin, D.: Spatial temporal graph convolutional networks for
  skeleton-based action recognition. CoRR  \textbf{abs/1801.07455} (2018),
  \url{http://arxiv.org/abs/1801.07455}

\end{thebibliography}
\end{document}